\documentclass[sigconf]{acmart}

\usepackage{bbm}
\usepackage{multirow}
\usepackage{mathtools}
\usepackage{stfloats}
\usepackage{balance}

\newcommand{\ours}{\textsc{FuseDiff}}

\AtBeginDocument{%
  }

\copyrightyear{2026}
\acmYear{2026}
\setcopyright{cc}
\setcctype{by}
\acmConference[KDD '26]{Proceedings of the 32nd ACM SIGKDD Conference on Knowledge Discovery and Data Mining V.2}{August 09--13, 2026}{Jeju Island, Republic of Korea}
\acmBooktitle{Proceedings of the 32nd ACM SIGKDD Conference on Knowledge Discovery and Data Mining V.2 (KDD '26), August 09--13, 2026, Jeju Island, Republic of Korea}
\acmDOI{10.1145/3770855.3819050}
\acmISBN{979-8-4007-2259-2/2026/08}

\begin{document}

\title{FuseDiff: Symmetry-Preserving Joint Diffusion for Dual-Target Structure-Based Drug Design}

\author{Jianliang Wu}
\orcid{0009-0008-4575-6748}
\affiliation{%
  \department{School of Computer Science and Engineering}
  \institution{Sun Yat-sen University}
  \city{Guangzhou}
  \country{China}
}
\email{wujliang23@mail2.sysu.edu.cn}

\author{Anjie Qiao}
\orcid{0009-0008-6406-5459}
\affiliation{%
  \department{School of Computer Science and Engineering}
  \institution{Sun Yat-sen University}
  \city{Guangzhou}
  \country{China}
}
\email{qiaoanj@mail2.sysu.edu.cn}

\author{Zhen Wang}
\orcid{0000-0002-8140-8782}
\authornote{Corresponding author.}
\affiliation{%
  \department{School of Computer Science and Engineering}
  \department{Guangdong Province Key Laboratory of Computational Science}
  \institution{Sun Yat-sen University}
  \city{Guangzhou}
  \country{China}}
\email{wangzh665@mail.sysu.edu.cn}

\author{Zhewei Wei}
\orcid{0000-0003-3620-5086}
\affiliation{%
 \department{Gaoling School of Artificial Intelligence}
 \institution{Renmin University of China}
 \city{Beijing}
 \country{China}}
\email{zhewei@ruc.edu.cn}

\author{Sheng Chen}
\orcid{0000-0003-1428-6778}
\affiliation{%
  \department{School of Life Science}
  \institution{Tsinghua University}
  \city{Beijing}
  \country{China}}
\email{chenshengsemail@mail.tsinghua.edu.cn}

\renewcommand{\shortauthors}{Jianliang Wu, Anjie Qiao, Zhen Wang, Zhewei Wei, \& Sheng Chen}

\begin{abstract}
Dual-target structure-based drug design aims to generate a single ligand together with two pocket-specific binding poses, each compatible with a corresponding target pocket, enabling polypharmacological therapies with improved efficacy and reduced resistance.
Existing approaches typically rely on staged pipelines, which either decouple the two poses via conditional-independence assumptions or enforce overly rigid correlations, and therefore fail to jointly generate two target-specific binding modes.
To address this, we propose \ours{}, an end-to-end diffusion model that jointly generates a ligand molecular graph and two pocket-specific binding poses conditioned on both pockets.
\ours{} features a message-passing backbone with Dual-target Local Context Fusion (DLCF), which fuses each ligand atom's local context from both pockets to enable expressive joint modeling while preserving the desired symmetries.
Together with explicit bond generation, \ours{} enforces topological consistency across the two poses under a shared graph while allowing target-specific geometric adaptation in each pocket.
To support principled training and evaluation, we derive a dual-target training set and use an independent held-out test set for evaluation.
Experiments on the benchmark and real-world dual-target pocket pairs show that \ours{} achieves state-of-the-art docking performance and enables the first systematic assessment of dual-target pose quality prior to docking-based pose search.
\end{abstract}




\maketitle

\section{Introduction}
\label{sec:intro}
Structure-based drug design (SBDD) exploits the three-dimensional structure of a biological target to design ligands with high affinity and selectivity for a given binding site~\cite{Anderson2003}.
While traditional SBDD has largely centered on a single target, the growing demand for polypharmacological agents that improve therapeutic efficacy and delay drug resistance~\cite{KABIR2022106055,YE2023188866} has increasingly motivated dual-target drug design.
Accordingly, \textbf{dual-target SBDD} seeks to design a single ligand together with \emph{two target-specific binding poses}, each compatible with its corresponding pocket.

With this goal, dual-target SBDD can be cast as conditional density estimation--one of the most natural and general formulations--namely learning $p(G, X_1, X_2 \mid P_1, P_2)$, where $P_1, P_2$ denote the two target pockets, $G$ denotes the ligand molecular graph, and $X_1, X_2$ denote the corresponding binding poses.
Intuitively, a sample from this distribution corresponds to a complete dual-target design proposal: one ligand identity, represented by a shared molecular graph $G$, together with two target-specific binding poses $X_1$ and $X_2$ that describe how the same ligand can be placed in the two pockets. This formulation explicitly couples molecular identity and dual-pocket geometry within a single conditional generative problem.
In contrast to search-based approaches that explore chemical space via multi-objective scoring~\cite{combimots,aixfuse}, this formulation directly supports de novo design and produces target-specific binding modes end-to-end, rather than treating pose recovery as a downstream docking step.
Another common shortcut is to reuse single-target generative models sequentially (or to combine their signals); when the first-stage sampling cannot account for the second pocket, the overall procedure can degrade into a nested-loop search over candidates, incurring substantial computational overhead.

Therefore, dedicated dual-target methods are needed.
Existing methods often adopt a two-stage pipeline: they first generate 2D molecular graphs (or SMILES) conditioned on both targets using chemical language models or synthesis-aware search, and then recover 3D binding poses via post-hoc optimization or separate single-target pose predictors~\cite{SUN2020112025}.
This design amounts to modeling $p(G \mid P_1,P_2)$ and then inferring $X_1$ and $X_2$ separately given $(G,P_1)$ and $(G,P_2)$, which largely decouples pose generation across targets and leaves little room to model cross-target coupling at the pose level.
In practice, the noise in paired pose observations (i.e., their experimental structures) tends to be correlated, making this conditional independence assumption restrictive.
Moreover, the staged procedure is prone to error accumulation.

To jointly generate ligand atom types and binding poses, \citet{dualdiff} reprogram a pretrained single-target diffusion model by composing predicted scores conditioned on each pocket.
To make this strategy viable, it requires explicit geometric alignment of the two pockets via a rigid transformation $\mathcal{T}$ estimated by probe-ligand docking.
The resulting sampling effectively targets a composed conditional $p(G, X \mid P_1, \mathcal{T} P_2)$, implicitly assuming the two target-specific poses are compatible under a single rigid transformation, which mismatches real dual-target pose distributions.
Consequently, the raw sampled poses can deviate substantially from the true binding modes and require docking-based conformational search to recover two poses, partially degenerating back to a two-stage pipeline.

In this work, we aim to learn $p(G, X_1, X_2 \mid P_1, P_2)$ in a genuinely end-to-end manner, while avoiding the restriction/assumption above to improve expressiveness.
To this end, we identify two fundamental obstacles.
First, dual-target SBDD datasets that explicitly support learning this density are scarce, as such data must jointly specify the ligand identity and paired binding poses across targets.
Second, jointly generating target-specific poses together with their shared molecular graph requires the model to introduce appropriate cross-target coupling.
However, modeling such coupling raises difficulty: overly strong coupling can violate desired symmetries (e.g., the model should respect rigid motions applied to each pocket-ligand complex), or even constrain the joint variability (e.g., collapsing them to be related by a single rigid motion); whereas insufficient coupling can yield poses that are mutually inconsistent with respect to a shared 2D molecular graph.

To address data scarcity, we derive a dual-target training set from BindingNet v2~\cite{bindingnetv2Zhu2025} with paired ligand poses across targets, and use the full DualDiff benchmark~\cite{dualdiff} as an independent held-out test set, enabling principled training and evaluation of dual-target SBDD methods.
To address modeling difficulty, we formalize the symmetries and expressiveness requirements for joint modeling.
We then present \textbf{\ours{}}, a conditional diffusion model that directly targets $p(G, X_1, X_2 \mid P_1, P_2)$ and satisfies these requirements by design.
At its core, \ours{} employs a message-passing backbone with \emph{Dual-target Local Context Fusion (DLCF)}, which allows each ligand atom to fuse the local context from both pockets, to achieve appropriate cross-target coupling.
Meanwhile, we explicitly model the bonds of the ligand to enhance topological consistency across both poses; without this, we consistently observe a failure mode in our pilot and ablation studies, where generated poses correspond to incompatible molecular graphs.
We train and evaluate \ours{} on the curated datasets and on real-world dual-target pocket pairs.
In both settings, \ours{} establishes the first benchmark for dual-target pose quality prior to docking-based pose search and achieves state-of-the-art performance under standard docking-based evaluation.

We summarize our main contributions as follows:
\begin{itemize}
  \item To our knowledge, we introduce the first end-to-end conditional diffusion model for de novo dual-target SBDD that directly targets the density $p(G, X_1, X_2 \mid P_1, P_2)$ while satisfying all the desired symmetries and remaining expressive enough to model the joint density.
  \item We derive a dual-target SBDD training dataset with paired poses for the same ligand across targets, enabling principled training of joint ligand-pose generation.
  \item We propose a message-passing backbone with \emph{Dual-target Local Context Fusion (DLCF)}, which enables suitable cross-target coupling and naturally generalizes to multi-target generative modeling.
  \item We establish the first benchmark for evaluating dual-target pose quality prior to docking-based pose search, and demonstrate state-of-the-art performance under standard docking-based metrics (Vina Dock) on both constructed and real-world pocket pairs.
\end{itemize}

\section{Related Work}\label{sec:related work}
\paragraph{Structure-Based Drug Design}
Recent SBDD methods formulate ligand design as 3D generative modeling conditioned on the pocket’s 3D structure~\cite{survey1Powers2023, survey2BAI2024104024}.
Early deep-learning approaches represented complexes on voxel grids~\cite{ligan,3dsbdd}, which limits spatial resolution due to discretization and is not SE(3)-equivariant by construction.
Continuous-coordinate methods leverage equivariant geometric GNNs~\cite{egnn} to construct ligands directly in 3D~\cite{liu2022graphbp,peng2022pocket2mol}; however, autoregressive assembly is sequential and can be sampling-inefficient. 
Diffusion models~\cite{ddpm} instead perform parallel denoising for one-shot generation~\cite{cbgbench,diffsbdd}; by jointly modeling continuous atom coordinates and discrete atom types, diffusion-based methods enable pose-consistent de novo design~\cite{targetdiff,anjiebib,qiao20253d}.
Recent evaluations further corroborate the strong performance of diffusion-based SBDD on structure-conditioned generation~\cite{cbgbench}. 
Nevertheless, principled dual-target extensions of such models remain limited.

\paragraph{Protein–ligand complex datasets}\label{relate:dataset}
Recent ML-based SBDD relies on protein-ligand complex datasets that provide 3D complex poses together with binding affinity data for supervised training and evaluation.
While PDB archives experimentally determined macromolecular structures~\cite{PDB}, curated resources such as PDBbind~\cite{pdbbind1Wang2004, pdbbind2Wang2005} and Binding MOAD~\cite{Hu2005BindingMOAD, BindingMOADimprovement} pair complex structures with literature-derived affinity measurements, and PDB-wide collections further expand affinity coverage~\cite{PDB-wide}. 
Because experimentally derived datasets remain limited in scale and diversity, larger pose collections are built via modeling: docking-based datasets such as CrossDocked2020 primarily augment pose diversity for docking pipelines~\cite{crossdocked2020}, whereas template-based datasets such as BindingNet v2 expand the coverage of experimentally annotated ligand–target pairs~\cite{bindingnetv2Zhu2025}. Dual-target resources are scarce; the DualDiff benchmark curates target pairs from synergistic drug combinations but provides distinct reference ligands per pocket, making it primarily an evaluation set~\cite{dualdiff}. Accordingly, we construct our training set from BindingNet v2 and reserve DualDiff dataset as an independent held-out evaluation set.

\paragraph{Dual-target ligand generation}
Dual-target ligand generation aims to design single molecules that can simultaneously engage a given pair of biological targets.
Most approaches generate molecules as SMILES or 2D graphs and may obtain docking poses via post-processing.
Representative approaches include substructure-based generators that construct molecular graphs from learned or predefined fragments~\cite{jin18a,jin20b}, as well as target-pair--conditioned chemical language models fine-tuned on ligands associated with the target pair~\cite{Isigkeit2024}.
Optimization-driven methods further incorporate multi-objective rewards, ranging from adversarial/RL generators~\cite{dlgn} to RL for polypharmacology~\cite{Munson2024polygon}. 
Complementarily, Pareto-front search explores trade-offs via Pareto MCTS, either in general chemical space~\cite{Yang2024paretodrug} or over synthesizable fragment libraries~\cite{combimots,aixfuse}. 
Structure-based dual-target ligand generation is less explored: \citet{dualdiff} align two pockets in 3D and reprogram pretrained single-target diffusion models for dual-target sampling, producing a ligand together with a 3D conformation.
In contrast, our work focuses on end-to-end joint generation of a shared molecular graph with two pocket-specific binding poses conditioned on the origin pocket pair.

\section{Methodology}
\label{sec:method}

Having formulated dual-target SBDD as learning the conditional density $p(G, X_1, X_2 \mid P_1, P_2)$, we begin by specifying the problem setting and notation in Sec.~\ref{subsec:Problem Definition}. 
Sec.~\ref{subsec:datasetconstruction} introduces the datasets used for training and evaluation. 
We elaborate on our proposed model \ours{} in Sec.~\ref{subsec:fusediff} and its core module, Dual-target Local Context Fusion (DLCF), in Sec.~\ref{sec:dlcf}. 
Finally, Sec.~\ref{subsec:analysis} formalizes the above intuition as requirements \textbf{R1--R4} and analyzes why \ours{} satisfies them.

\subsection{Problem Definition}
\label{subsec:Problem Definition}

The probabilistic formulation in Sec.~\ref{sec:intro} can be understood as defining the distribution of complete dual-target design objects. In this setting, each generated sample should contain not only the chemical identity of the ligand, but also two pocket-specific 3D poses that are tied to the same ligand graph.
Given a pair of protein binding sites $(\mathcal{P}_1, \mathcal{P}_2)$, each pocket $\mathcal{P}_k$ is represented as a set of $N_{P_k}$ atoms 
$\mathcal{P}_k = \{(x_{P_k}^{(i)}, v_{P_k}^{(i)})\}_{i=1}^{N_{P_k}},k\in \{1,2\}$
where $x_{P_k}^{(i)}\in\mathbb{R}^3$ denotes the 3D coordinates of the $i$-th pocket atom,
and $v_{P_k}^{(i)}\in \{0,1\}^{N_f}$ denotes its feature (e.g., element type and amino-acid type).
Our goal is to generate a ligand molecular graph 
$\mathcal{G}=(\mathcal{V}, \mathcal{B}), \mathcal{V}=\{v_i\}_{i=1}^{N_M}, \mathcal{B}=\{b_{ij}\}_{i,j=1}^{N_M}$ together with its two target-specific binding conformations $\mathcal{X}_1=\{ x_{1,i}\}_{i=1}^{N_M}$ and $\mathcal{X}_2=\{ x_{2,i} \}_{i=1}^{N_M}$, 
conditioned on both pockets, 
where $x_{1,i}\in\mathbb{R}^3$ denotes the 3D coordinate of the $i$-th ligand atom in the $\mathcal{P}_1$-specific binding conformation $\mathcal{X}_1$ (analogously for $\mathcal{X}_2$), 
$v_i\in\{0,1\}^{N_v}$ denotes its atom type,
and $b_{ij}\in\{0,1\}^{N_b}$ denotes the bond type between atoms $i$ and $j$, with $N_v$ and $N_b$ are the numbers of atom and bond types, respectively.
In discussing density $p(G, X_1, X_2 \mid P_1, P_2)$, we use $P_k$, $G$, and $X_k$ to denote both random variables and their corresponding realizations $\mathcal{P}_k$, $\mathcal{G}$, and $\mathcal{X}_k$, when there is no ambiguity.
An ideal generative model should satisfy the following requirements:

This formulation entails four basic desiderata for dual-target generation. 
First, the two targets should be treated symmetrically: exchanging the order of the pockets should only exchange the corresponding poses, without changing the distribution of the shared ligand graph. 
Second, each pocket--ligand complex should respect SE(3) equivariance. 
Third, the two poses should be jointly modeled, i.e., $X_1$ and $X_2$ should not factorize given $P_1$, $P_2$, and the shared graph $G$. 
Fourth, the joint distribution should not impose overly restrictive geometric constraints, such as requiring one pose to be obtained from the other by a rigid transformation. 
We formalize these desiderata as \textbf{R1--R4} in Sec.~\ref{subsec:analysis} after introducing the model architecture.

\subsection{Dual-target Datasets}
\label{subsec:datasetconstruction}
To support end-to-end learning of the density $p(G, X_1, X_2 \mid P_1, P_2)$, we require tuples in the form $(G, X_1, X_2, P_1, P_2)$. Since no public resource natively provides such tuples at scale, we derive the training data ourselves and use the DualDiff benchmark (DDF) for independent evaluation.

\paragraph{Training set BN2-DT}
We build a dual-target training set from the high confidence subset of BindingNet v2 and term it \emph{BindingNet v2--Dual-Target (BN2-DT)}.
BindingNet v2 provides single-target complexes, which we view as entries $(G, X, P)$. A dual-target instance $(G, X_1, X_2, P_1, P_2)$ is included if and only if two such entries share graph-isomorphic ligands (identical atom types and bond orders) and correspond to two distinct pockets/targets. Specifically, for two entries, namely $(G_1, X_1, P_1)$ and $(G_2, X_2, P_2)$, if $G_1\!\equiv\!G_2\!\wedge\!P_1\!\not\approx\!P_2$, we derive a dual-target tuple $(G\!=\!G_1\!=\!G_2, X_1, X_2, P_1, P_2)$, where $P_1\!\not\approx\!P_2$ denotes that the two pockets are distinct (Appendix~\ref{app:dataset}). 
We aggregate all eligible pairs with basic cleaning and deduplication, resulting in $58{,}058$ dual-target training tuples with $801$ unique protein targets and $21{,}721$ unique ligands. Full construction details and basic statistics are deferred to Appendix~\ref{app:dataset}.

\paragraph{Test set DDF}
For an independent evaluation, we adopt the dual-target benchmark introduced by \citet{dualdiff}, which curates \emph{paired targets} from synergistic drug combinations and provides a \emph{reference ligand} for each target.
Consequently, each record can be treated as $(G_1, X_1, P_1,\, G_2, X_2, P_2)$, where typically $G_1 \neq G_2$ since the two pockets have distinct reference ligands.
We refer to this benchmark as the \emph{\textsc{DualDiff} test set (DDF)} and use it solely as an independent held-out test set.

\begin{figure*}[t]
\centering
\includegraphics[width=0.925\textwidth]{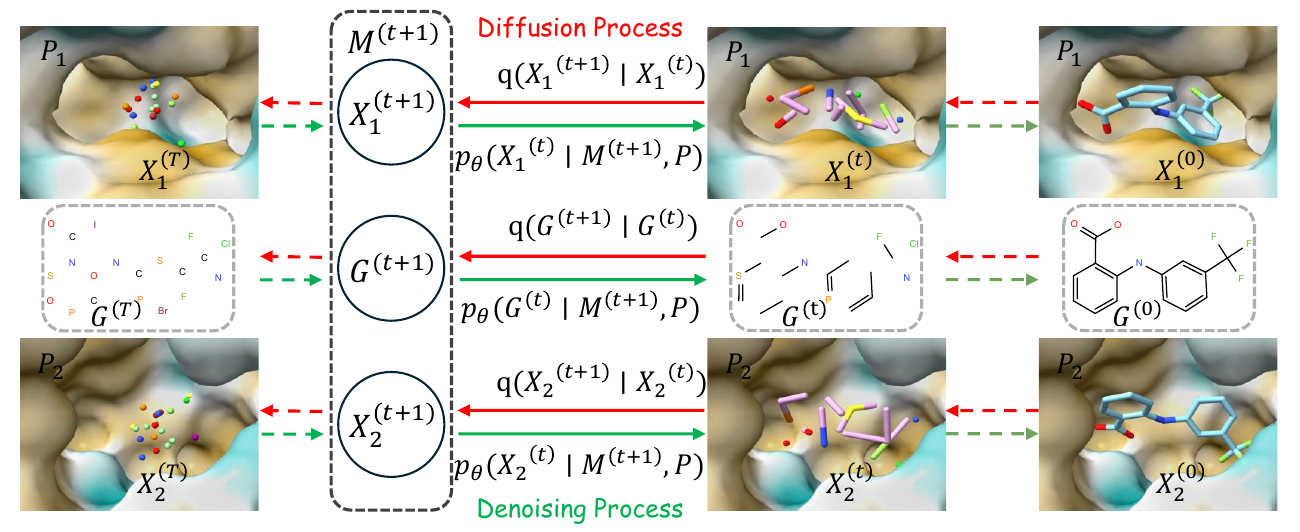}
\caption{\textbf{Overview of \ours.}
\ours{} defines a conditional diffusion model over $M=(G,X_1,X_2)$ given a pocket pair $P=(P_1,P_2)$, where the two target-specific complexes share the same molecular graph $G$ while maintaining separate poses in $P_1$ and $P_2$.
A fixed forward (diffusion) process $q$ (red arrows) corrupts $(G^{(0)},X_1^{(0)},X_2^{(0)})$ into $(G^{(T)},X_1^{(T)},X_2^{(T)})$ for training.
For generation, the learned reverse (denoising) process $p_\theta$ (green arrows) starts from $M^{(T)}\!\sim\! p_{\text{base}}(M^{(T)}\mid P)$ and iteratively samples $M^{(t-1)}$ using a denoiser $f_\theta$.
At step $t$, $f_\theta$ takes $(G^{(t)},X_1^{(t)},X_2^{(t)})$, $t$, and $P$ as input and predicts $(\widehat{X}_1^{(0)},\widehat{X}_2^{(0)},\widehat{V}^{(0)},\widehat{B}^{(0)})$ to parameterize the reverse transition.
}
\label{fig:overview}
\end{figure*}

\subsection{Diffusion Models for dual-target SBDD}
\label{subsec:fusediff}

\paragraph{Overview of \ours.}\label{overview}

As shown in Fig.~\ref{fig:overview}, \ours{} is a diffusion model that consists of a fixed forward (diffusion) process and a trainable reverse (denoising) process.
For notational convenience, we denote the complete dual-target design object as $M=(G,X_1,X_2)$ and the conditioning pocket pair as $P=(P_1,P_2)$. 
Under this notation, learning $p(G,X_1,X_2 \mid P_1,P_2)$ is equivalent to learning the conditional generative distribution $p_\theta(M \mid P)$. \ours{} implements this formulation by applying diffusion to the entire object $M$: the forward process corrupts the shared molecular graph $G$ and both target-specific poses $X_1$ and $X_2$, while the reverse process jointly denoises all components conditioned on the two pockets. Consequently, one reverse sampling trajectory directly produces a shared ligand graph together with its two pocket-specific binding poses, without invoking a separate pose-recovery step.
Formally, the conditional distribution estimated by \ours{} is expressed as
$p_\theta(M^{(0)}\mid P)=\int p_\theta(M^{(0:T)}\mid P)\mathrm{d}M^{(1:T)}$, where $M^{(t)}$ for $t=1,\dots,T$ is a sequence of latent variables with the same dimensionality as the data $M^{(0)}$.
The forward process gradually corrupts data until reaching a stationary base distribution.
The reverse process performs step-wise denoising with a denoiser, recovering samples back towards the data distribution.

To enhance \emph{topology consistency}: generating different $X_1$ and $X_2$ while keeping both consistent with the generated $G$, \ours{} generates $G$---including both atom types $V$ and bond types $B$ as defined in Sec.~\ref{subsec:Problem Definition}---together with $(X_1, X_2)$, thus enforcing a single shared molecular graph during joint generation.
Without explicitly generating $B$, bonds are inferred post hoc for each pose independently.
In our ablation study (Sec.~\ref{subsec:ablation}), removing explicit bond-type modeling breaks topology consistency and prevents generating plausible samples.

The diffusion denoiser takes the noisy variables $M^{(t)}=(V^{(t)},\allowbreak B^{(t)},\allowbreak X_1^{(t)},\allowbreak X_2^{(t)})$, the timestep $t$, and the pair of conditioning pockets $P$, as input, and predicts all components of $M^{(0)}$ jointly.
To achieve suitable cross-target coupling, our denoiser relies on Dual-target Local Context Fusion (DLCF) elaborated in Sec.~\ref{sec:dlcf}.

\paragraph{Diffusion Process.}\label{forwardprocess}
Following the convention in single-target SBDD methods~\cite{targetdiff}, we apply Gaussian diffusion to the continuous coordinates $(X_1, X_2)$ and categorical diffusion for discrete graph attributes $G=(V,B)$, including atom and bond types.
The $t$-step forward marginals factorize as:
\begin{equation}
q(M^{(t)}\mid M^{(0)})=q(G^{(t)}\mid G^{(0)})q(X_{1}^{(t)}\mid X_{1}^{(0)})q(X_{2}^{(t)}\mid X_{2}^{(0)}),
\label{eq:fwdmargin}
\end{equation}
where the noise is injected independently for each component ($X_1, X_2$, and $G$), and within each component independently across atoms and edges.
Specifically, the categorical transition is defined by $q(G^{(t)} \mid  G^{(0)})=q(V^{(t)}\mid  V^{(0)})\,q(B^{(t)}\mid  B^{(0)})$ with
$q(V^{(t)}\mid  V^{(0)})=\prod_{i=1}^n \mathcal{C}\!\Big(v_i^{(t)} \mid  \bar\alpha_t v_i^{(0)} + (1-\bar\alpha_t)\mathrm{Unif}([N_v])\Big)$ and
$q(B^{(t)}\mid  B^{(0)})=\prod_{i<j} \mathcal{C}\!\Big(b_{ij}^{(t)} \mid  \bar\alpha_t b_{ij}^{(0)} + (1-\bar\alpha_t)\mathbbm{1}_{\textit{none-type}}\Big)$.
where $\mathrm{Unif}([N_v])$ is the uniform categorical over $[N_v]=\{1,\ldots,N_v\}$, and $\mathbbm{1}_{\textit{none-type}}$ is the one-hot categorical concentrated on the ``none-type''.
For $k\in\{1,2\}$, we apply the i.i.d.\ Gaussian noise per atom:
$q(X_k^{(t)}\mid  X_k^{(0)})=\prod_{i=1}^n \mathcal{N}\!\Big(x_{k,i}^{(t)} \mid  \sqrt{\bar\alpha_t}\,x_{k,i}^{(0)}, (1-\bar\alpha_t)\mathbf I\Big)$.
Here $\{\beta_t\}_{t=1}^T$ is a predefined noise scheduler, and we denote $\alpha_t=1-\beta_t$ and $\bar\alpha_t=\prod_{s=1}^t \alpha_s$ for the cumulative signal coefficient. 
We follow~\citet{moldiff} for the noise schedule $\{\bar{\alpha}_t\}_{t=1}^T$, including a bond-first corruption that quickly absorbs bonds to the \textit{none-type}.
As $t\!\to\!T$, the forward marginals approach a simple terminal prior, which we take as the base distribution to initialize reverse sampling: $p_{\text{base}}(M^{(T)}\mid P)
=\mathcal{N}(X_1^{(T)};0,\mathbf I)\,\allowbreak
\mathcal{N}(X_2^{(T)};0,\mathbf I)\,\allowbreak
\mathcal{C}(V^{(T)};\mathrm{Unif}(N_v))\,\allowbreak
\mathcal{C}(B^{(T)};\mathbbm{1}_{\textit{none-type}})$.

\newpage
\paragraph{Parameterization of the denoising process.}\label{para:reverseprocess}
Starting from the base prior $p_{\text{base}}(M^{(T)}\allowbreak\mid  P)$, we define a Markov chain
$p_\theta(M^{(0:T)}\allowbreak\mid  P)=p_{\text{base}}(M^{(T)}\allowbreak\mid  P)\prod_{t=1}^T p_\theta(M^{(t-1)}\allowbreak\mid  M^{(t)},\allowbreak P)$.
Following the factorization of the diffusion process (Eq.~\ref{eq:fwdmargin}), we factorize the reverse transition as 
$
p_\theta(M^{(t-1)}\allowbreak\mid  M^{(t)},\allowbreak P)
=\Big(\prod_{k=1}^2 p_\theta(X_k^{(t-1)}\allowbreak\mid  M^{(t)},\allowbreak P)\Big)\,
p_\theta(V^{(t-1)}\allowbreak\mid  M^{(t)},\allowbreak P)\,
p_\theta(B^{(t-1)}\allowbreak\mid  M^{(t)},\allowbreak P)
$, where all these terms are approximated by substituting the denoised components into the one-step posteriors~\cite{ddpm,moldiff}: $p_\theta(X_k^{(t-1)}\!\mid M^{(t)},\allowbreak P)=q(X_k^{(t-1)}\!\mid X_k^{(t)},\allowbreak\widehat{X}_k^0)$,
$p_\theta(V^{(t-1)}\!\mid M^{(t)},\allowbreak P)=q(V^{(t-1)}\!\mid V^{(t)},\allowbreak\widehat{V}^0)$, and
$p_\theta(B^{(t-1)}\!\mid M^{(t)},\allowbreak P)=q(B^{(t-1)}\!\mid B^{(t)},\allowbreak\widehat{B}^0)$.
All these denoised components are jointly predicted by a single denoiser $f_\theta$:
\begin{equation}
(\widehat{X}_1^{(0)},\widehat{X}_2^{(0)},\widehat{V}^{(0)},\widehat{B}^{(0)})=f_\theta(M^{(t)},t,P).
\label{eq:denoiesr}
\end{equation}

\paragraph{Optimization.}
At each training iteration, we sample a training tuple $(M, P)$ and draw $t\sim\mathrm{Unif}(\{1,\ldots,T\})$.
We then obtain $M^{(t)}\sim q(M^{(t)}\!\mid M^{(0)}=M)$ by Eq.~\ref{eq:fwdmargin}, and make predictions by Eq.~\ref{eq:denoiesr}.
%$(\widehat{X}_1^{(0)},\widehat{X}_2^{(0)},\widehat{V}^{(0)},\widehat{B}^{(0)})=f_\theta(M^{(t)},t,P)$.
The clean sample and these predictions together define a weighted denoising objective to optimize $f_\theta$:
\begin{equation*}
\resizebox{\linewidth}{!}{$
\mathcal{L}_{t-1}\!=\!
\lambda\!\left(\mathcal{L}^{\text{pos}_1}_{t-1}\!+\!\mathcal{L}^{\text{pos}_2}_{t-1}\right)\!+\!
\lambda_v\,\mathcal{L}^{\text{a-type}}_{t-1}\!+\!
\lambda_b\,\mathcal{L}^{\text{b-type}}_{t-1}\!+\!
\lambda_{bl}\,\mathcal{L}^{\text{b-len}}_{t-1}.
$}
\label{eq:loss_total}
\end{equation*}
Here $\mathcal{L}^{\text{pos}_k}_{t-1}$ is the MSE between $\widehat{X}_k^{(0)}$ and $X_k^{(0)}$.
$\mathcal{L}^{\text{a-type}}_{t-1}$ and $\mathcal{L}^{\text{b-type}}_{t-1}$ are KL divergences that match the predicted one-step posteriors parameterized by $\widehat{V}^{(0)}$ and $\widehat{B}^{(0)}$ to its counterpart induced by ground-truth $V^{(0)}$ and $B^{(0)}$.
We additionally include a MSE loss $\mathcal{L}^{\text{b-len}}_{t-1}$ for bond-length over bonded pairs, regressing interatomic distances to the ground truth to enforce plausible covalent lengths.
The weights $\lambda,\lambda_v,\lambda_b,\lambda_{bl}$ are nonnegative hyperparameters controlling the relative contributions of each term.

\begin{figure*}[t]
\centering
\includegraphics[width=\textwidth]{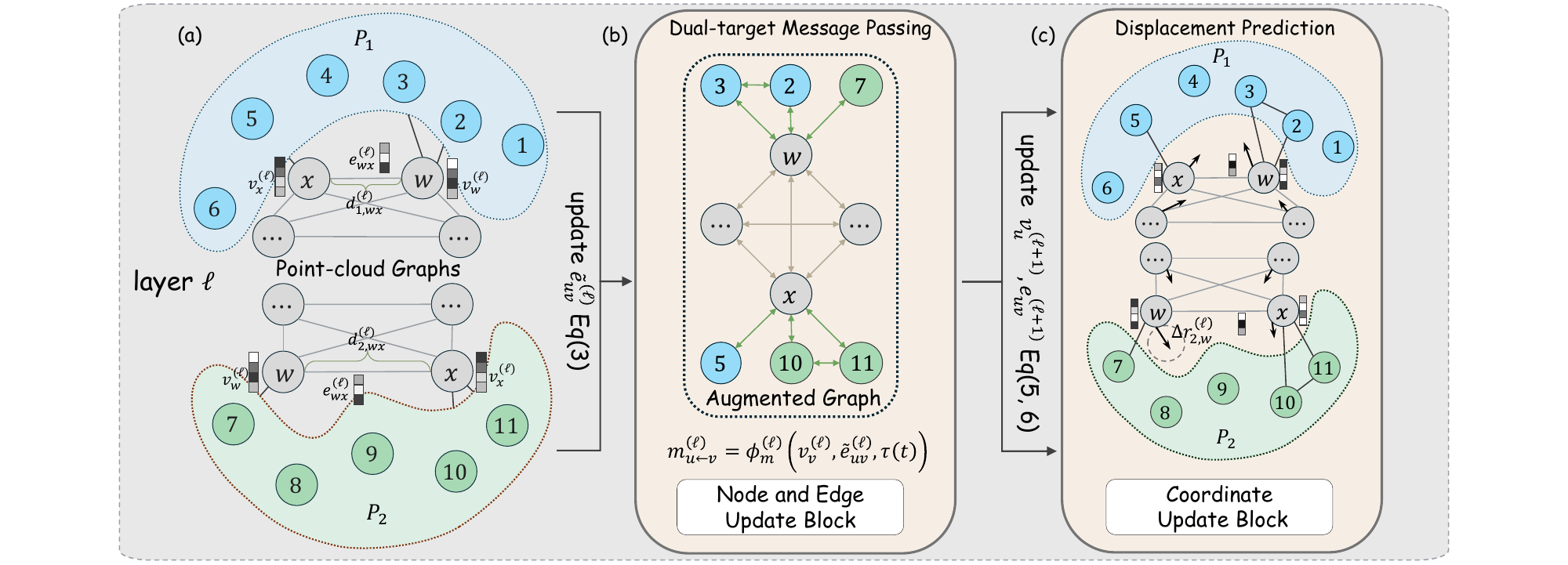}
\caption{\textbf{Dual-target Local Context Fusion (DLCF) within one denoising layer.}
(a) Two pocket--ligand point-cloud graphs for $P_1$ and $P_2$, each containing a fully-connected ligand subgraph.
(b) The \emph{augmented graph} induced from the two point-cloud graphs, on which dual-target message passing updates node/edge embeddings.
(c) Coordinate update block that \emph{separately} updates ligand node coordinates in the two point-cloud graphs (with pocket atoms fixed).}
\Description{dlcf}
\label{fig:dlcf}
\end{figure*}

\subsection{Dual-target Local Context Fusion (DLCF)}
\label{sec:dlcf}

We parameterize the denoiser $f_\theta$ as an $L$-layer MPNN~\cite{mpnn} that is SE(3)-equivariant under independent rigid motions of each pocket--ligand complex.
To predict the reverse transition in Eq.~\ref{eq:denoiesr}, $f_\theta$ takes the current noisy complexes at denoising step $t$ as input.
In each layer $\ell$, $f_\theta$ dynamically constructs two pocket--ligand point-cloud graphs and fuses them into a single augmented graph, then updates node/edge representations via message passing on the augmented graph (Eq.~\ref{eq:cal_e_tilde}--\ref{eq:mp_edge_update}), thereby injecting local context from both pockets into shared ligand atom/bond embeddings, and finally updates ligand coordinates in the two point-cloud graphs separately (Eq.~\ref{eq:coord_update_masked}), as illustrated in Fig.~\ref{fig:dlcf}.

\subsubsection{Dual-target graph setup.}
\label{subsubsec:graphsetup}

At the $\ell$-th MPNN layer ($\ell\in\{1,\allowbreak\ldots,\allowbreak L\}$), we denote the current ligand--pocket complexes by $R_k^{(\ell)}\coloneqq \{\mathbf r^{(\ell)}_{k,u}\}_{u\in V}$ for $k\in\{1,2\}$.
Here, the subscript $u$ indexes the atom, and thus $\mathbf r^{(\ell)}_{k,u}\in\mathbb{R}^3$ denotes the coordinate of the atom $u$ in the $k$-th complex at layer $\ell$.
When $u$ corresponds to a pocket's atom, $\mathbf r^{(\ell)}_{k,u}$ always takes value from the condition $P_k$ across the layers.
When $u$ corresponds to the ligand's atom, we initialize $\mathbf r^{(1)}_{k,u}$ by the input noisy coordinate $X_{k,u}^{(t)}$ and update it across the $L$ layers.
We construct a protein--ligand point-cloud graph $G_k^{(\ell)}$.
The nodes of $G_k^{(\ell)}$ consist of ligand atoms and pocket atoms in $P_k$.
Within $G_k^{(\ell)}$, ligand nodes are fully connected, while all remaining edges are constructed by 3D $k$-nearest neighbors in terms of $R_{k}^{(l)}$ following~\citet{targetdiff}.
We use $\mathcal{N}_k^{(\ell)}(u)$ to denote the neighbors of node $u$ in $G_k^{(\ell)}$.

We then build an \emph{augmented graph} $G_{\mathrm{aug}}^{(\ell)}$ by fusing the local neighborhoods of $G_1^{(\ell)}$ and $G_2^{(\ell)}$ for dual-target message passing.
Since $G_1^{(\ell)}$ and $G_2^{(\ell)}$ share the same ligand identities but have target-specific coordinates, we deduplicate ligand nodes and define the node set of $G_{\mathrm{aug}}^{(\ell)}$ as $V \cup P_1 \cup P_2$.
Here, $V$, $P_1$, and $P_2$ are overloaded to denote the disjoint node subsets corresponding to the ligand atoms and the pocket atoms from the two targets, respectively, when there is no ambiguity.
In $G_{\mathrm{aug}}^{(\ell)}$, ligand nodes remain fully connected and all intra-pocket edges are preserved.
Crucially, for each ligand node $u\in V$, its pocket-atom neighbors are inherited from \emph{both} $G_1^{(\ell)}$ and $G_2^{(\ell)}$.
Formally, let $\mathcal{N}^{(\ell)}(u)$ denote the neighbors of $u$ in $G_{\mathrm{aug}}^{(\ell)}$.
Accordingly, for a pocket node $u\in P_k$, $\mathcal{N}^{(\ell)}(u)=\mathcal{N}_k^{(\ell)}(u)$; for a ligand node $u\in V$, $\mathcal{N}^{(\ell)}(u)=\mathcal{N}_1^{(\ell)}(u)\cup\mathcal{N}_2^{(\ell)}(u)$ after deduplicating ligand nodes.
An example of this fusion procedure is illustrated in Fig.~\ref{fig:dlcf}~((a)$\rightarrow$(b)).

The $l$-th layer MPNN will perform message passing on $G_{\mathrm{aug}}^{(\ell)}$.
We denote current node and edge representations by $v^{(\ell)}_{u}\in\mathbb{R}^{d_v}$ and $e^{(\ell)}_{uv}\in\mathbb{R}^{d_e}$, respectively, with $\ell=1$ initialized from input atom/bond type embeddings together with the timestep embedding $\tau(t)$.

\subsubsection{Dual-target message-passing.}
\label{subsubsec:messagepassing}

For each edge $(u,v)$ in $G_{\mathrm{aug}}^{(\ell)}$,
we first compute a distance-aware edge embedding $\tilde e^{(\ell)}_{uv}$ by injecting the inter-atomic distance $d_{k,uv}^{(\ell)}=\|\mathbf{r}^{(\ell)}_{k,u}-\mathbf{r}^{(\ell)}_{k,v}\|_2$ into the edge representation $e^{(\ell)}_{uv}$:
% , where $\mathbf{r}^{(\ell)}_{k,u}\in\mathbb{R}^3$ denotes the absolute 3D coordinate of node $u$ in $G_k$
\begin{equation}
\tilde e^{(\ell)}_{uv}=
\begin{cases}
\phi_{d,v}^{(\ell)}\!\left(e^{(\ell)}_{uv},\, s_{uv}^{(\ell)},\, \delta_{uv}^{(\ell)}\right), & u,v\in V,\\[2pt]
\phi_{d,p}^{(\ell)}\!\left(e^{(\ell)}_{uv},\, d_{k,uv}^{(\ell)}\right), & u\in P_k \ \text{or}\ v\in P_k.
\end{cases}
\label{eq:cal_e_tilde}
\end{equation}
Here, note that $G_{\mathrm{aug}}^{(\ell)}$ contains no edges between $P_1$ and $P_2$, hence $k$ is uniquely determined in the second case.
Meanwhile, by letting $s_{uv}^{(\ell)}=d_{1,uv}^{(\ell)}+d_{2,uv}^{(\ell)}$ and $\delta_{uv}^{(\ell)}=|d_{1,uv}^{(\ell)}-d_{2,uv}^{(\ell)}|$, the formulation in the first case is invariant to swapping the two targets.
Distance-aware edge embeddings encode the required geometric cues, allowing us to perform node/edge updates on $G_{\mathrm{aug}}^{(\ell)}$ using only scalar distances without explicit 3D relative displacement vectors.
%Conditioned on the distance-aware edge embedding and node embedding,
Then, we compute a directed message from node $v$ to node $u$ via:
\begin{equation}
m^{(\ell)}_{u\leftarrow v}=\phi_m^{(\ell)}\!\left(
v^{(\ell)}_{v},\, \tilde e^{(\ell)}_{uv},\, \tau(t)\right).
\label{eq:mp_msg}
\end{equation}
The above transformations $\phi_{d,v}^{(\ell)}, \phi_{d,p}^{(\ell)}$, and $\phi_m^{(\ell)}$ are neural networks composed of different multilayer perceptrons (MLPs).
Finally, we update node representations $v^{(\ell+1)}_{u}$ by aggregating messages from $\mathcal{N}^{(\ell)}(u)$:
\begin{equation}
v^{(\ell+1)}_{u} = \mathrm{Linear}\!\left(v^{(\ell)}_{u}\right)\!+\!
\sum\limits_{v\in\mathcal{N}^{(\ell)}(u)} m^{(\ell)}_{u\leftarrow v}.
\label{eq:mp_node_update}
\end{equation}
This design allows each ligand node to incorporate pocket-local context from both targets, since $\mathcal{N}^{(\ell)}(u)$ contains atoms from both pockets.
Similarly, edge representations $e^{(\ell)}_{uv}$ are updated via message passing according to:

\begin{equation}
\resizebox{\linewidth}{!}{$
e^{(\ell+1)}_{uv}
=
\sum\limits_{\xi\in\{u,v,e\}}\mathrm{Linear}_\xi\!\left(h^{(\ell)}_{\xi}\right)
+
\sum\limits_{s\in\{u,v\}}
\sum\limits_{w\in\mathcal{N}^{(\ell)}(s)}
m^{(\ell)}_{s\leftarrow w}
$},
\label{eq:mp_edge_update}
\end{equation}
where $h^{(\ell)}_{u}\!=\!v^{(\ell)}_{u},
h^{(\ell)}_{v}\!=\!v^{(\ell)}_{v},
h^{(\ell)}_{e}\!=\!e^{(\ell)}_{uv}$.

Unlike node/edge representation updates that are performed on $G_{\mathrm{aug}}^{(\ell)}$, coordinate updates are performed separately in $G_k^{(\ell)},k\in\{1,2\}$.
Moreover, we update only the coordinates of the ligand atoms while keeping the coordinates of the pocket atoms fixed.
Specifically, for each $G_k^{(\ell)}$, we predict an SE(3)-equivariant displacement for every ligand node using the just-updated node/edge embeddings: $\mathbf{r}^{(\ell+1)}_{k,u}=\mathbf{r}^{(\ell)}_{k,u}+\Delta\mathbf{r}^{(\ell)}_{k,u}$,
where $\Delta\mathbf{r}^{(\ell)}_{k,u}$ is calculated as follows:
\begin{equation}
%\resizebox{\linewidth}{!}{$\displaystyle
\Delta\mathbf r^{(\ell)}_{k,u}
=\sum_{\mathclap{w\in\mathcal{N}_k(u)}}\!
\phi_r^{(\ell)}\!\bigl(
v^{(\ell+1)}_{u},\,
v^{(\ell+1)}_{w},\,
\tilde e^{(\ell+1)}_{uw},\,
\tau(t)\bigr)\,
\frac{r_{k,u}^{(\ell)}-r_{k,w}^{(\ell)}}{{d^{(\ell)}_{k,uw}}^2}.
\label{eq:coord_update_masked}
%}
\end{equation}
The update is SE(3)-equivariant because $\phi_r^{(\ell)}$ uses only invariant inputs and scales a relative vector.

Importantly, updating coordinates separately on $G_k^{(\ell)}$ does not imply single-pocket conditioning or a two-stage design. The displacement field depends on the newly updated node and edge representations that have already fused local contexts from both pockets.

\paragraph{Extension to multi-target.}
The above design extends naturally to $K$ targets by building $K$ pocket--ligand point-cloud graphs.
We fuse them into an augmented graph by starting from the fully-connected ligand subgraph, attaching to each ligand node $u$ its pocket-atom neighbors from all $K$ point-cloud graphs, and retaining all intra-pocket edges.
Node/edge embeddings are updated on this augmented graph as before, while ligand coordinates are updated independently in each point-cloud graph.

\subsection{Analysis}
\label{subsec:analysis}

We now identify four requirements \textbf{R1--R4} for generative modeling of $p(G, X_1, X_2 \mid P_1, P_2)$, which formalize the above modeling desiderata.

\begin{definition}[Requirements \textbf{R1--R4}]
The learned $p(G, X_1, X_2 \mid P_1, P_2)$ satisfies \textbf{R1 permutation invariance} if $\forall \mathcal{G}, \mathcal{P},\mathcal{P}',\mathcal{X},\mathcal{X}'$, $p(\mathcal{G}, \mathcal{X}, \mathcal{X}' \mid \mathcal{P}, \mathcal{P}') = p(\mathcal{G}, \mathcal{X}', \mathcal{X} \mid \mathcal{P}', \mathcal{P})$.
It satisfies \textbf{R2 SE(3) invariance} if $\forall g,g'\in\text{SE(3)}, p(\mathcal{G}, \mathcal{X}, \mathcal{X}' \mid \mathcal{P}, \mathcal{P}') = p(\mathcal{G}, \mathcal{T}_{g}\mathcal{X}, \mathcal{T}_{g'}\mathcal{X}' \mid \mathcal{T}_{g}\mathcal{P}, \mathcal{T}_{g'}\mathcal{P}')$.
It satisfies \textbf{R3 Non-factorization} if $X_1 \not\perp X_2 \mid P_1, P_2, G$.
It satisfies \textbf{R4 Non-restricted support} if its support is not restricted to $\{(\mathcal{X}_1,\mathcal{X}_2): \exists g\in\text{SE(3) } s.t., \mathcal{X}_1 = \mathcal{T}_g \mathcal{X}_2\}$.
\label{def:requirements}
\end{definition}

Based on these requirements, we check whether existing methods and our proposed \ours{} satisfy them.
Benefited from their decomposition of the target distribution and the adoption of geometric GNNs, most existing methods satisfy \textbf{R1\&R2}.
However, the benefit is not free: The molecule-then-poses strategy~\cite{SUN2020112025} breaks \textbf{R3}; and the pocket alignment preprocessing~\cite{dualdiff} breaks \textbf{R4}.

\begin{proposition}
\label{prop:r1r4}
\ours{} satisfies \textbf{R1--R4}.
\end{proposition}

Due to the limited space, we defer the proof to Appendix~\ref{app:theory}.

\section{Experiment}
\label{sec:exp}
Our experiments evaluate the chemical properties, binding affinity proxies and dual-target binding compatibility of generated ligands, verify the necessity of our key design through ablations, and further demonstrate real-world utility through a main-text Alzheimer's disease polypharmacology case on GSK3$\beta$--JNK3, with an additional ROR$\gamma$t--DHODH case provided in Appendix~\ref{app:ror-dhodh}.
Our code and data are available at \url{https://github.com/SYSUWuJL/FuseDiff}.

\subsection{Experimental Setup}
\paragraph{Dataset.}
We train \ours{} on our derived dual-target training set BN2-DT and evaluate on DDF (Sec.~\ref{subsec:datasetconstruction}).
For controlled comparison, we train a \textsc{TargetDiff} checkpoint on BindingNet v2 and use it as the pretrained single-target model for \textsc{TargetDiff}, \textsc{DualDiff}, and \textsc{CompDiff}.
This controls for differences in single-target pretraining data, so the results reflect the dual-target modeling design rather than the choice of pretraining dataset.
Appendix~\ref{app:targetdiff_bn2} shows that the BindingNet v2 checkpoint performs on par with the official CrossDocked2020 checkpoint on DDF, supporting its use as a common pretrained backbone for fair benchmarking.

\paragraph{Baselines.}
We compare \ours{} with the reference ligands in the test set and the ligands generated by several baselines. \textsc{TargetDiff}~\cite{targetdiff} is a diffusion-based method originally proposed for single-target de novo SBDD. \textsc{LinkerNet}~\cite{guan2023linkernet} is a diffusion-based model for co-designing molecular fragment poses and the linker. Following~\citet{dualdiff}, we apply \textsc{TargetDiff} and repurpose \textsc{LinkerNet} to the dual-target setting. \textsc{DualDiff} and \textsc{CompDiff}~\cite{dualdiff} are dual-target sampling methods that reprogram pretrained single-target diffusion models by aligning two binding pockets. 
We exclude search-based methods (e.g., \textsc{CombiMOTS}) that rely on MCTS with iterative oracle/docking evaluations over an external building-block space, as their compute-budgeted search setting is not directly comparable to our learned generative baselines.

\paragraph{Metrics.}
We evaluate generated ligands from two perspectives: \textbf{molecular properties} and \textbf{protein--ligand interactions}.

\textbf{molecular properties.}\quad
We adopt widely used metrics from prior work~\cite{cbgbench,targetdiff} and report (\romannumeral 1) \textbf{QED ($\uparrow$)}~\cite{qed} for drug-likeness;
(\romannumeral 2) \textbf{SA ($\uparrow$)}~\cite{sa} for synthetic accessibility;
(\romannumeral 3) \textbf{Diversity (Div.) ($\uparrow$)} as the average pairwise Tanimoto distance between molecular fingerprints;
(\romannumeral 4) \textbf{logP} measuring lipophilicity via the octanol--water partition coefficient (typically $[-0.4, 5.6]$ for drug-like compounds); and
(\romannumeral 5) \textbf{LPSK ($\uparrow$)}, the number of satisfied criteria in Lipinski's Rule-of-Five~\cite{LIPINSKI19973}, averaged over generated molecules.

\textbf{interactions.}\quad
For structure-based evaluation of protein--ligand interactions, we report widely used binding-affinity proxies computed by AutoDock Vina:
(\romannumeral 1) \textbf{Vina Score ($\downarrow$)}, the affinity of a given pose in the target pocket (scoring the given pose without optimization);
(\romannumeral 2) \textbf{Vina Min ($\downarrow$)}, the affinity after local optimization (energy minimization) in the pocket;
(\romannumeral 3) \textbf{Vina Dock ($\downarrow$)}, the best affinity after conformational search in the pocket.
Accordingly, \textbf{Vina Score} and \textbf{Vina Min} are both highly dependent on the input pose and evaluate poses prior to docking-based pose search, whereas \textbf{Vina Dock} is a docking-based score that performs global conformational search.
As Vina scores are biased by ligand size (larger molecules tend to score lower)~\cite{gao2025reframingsbdd,cbgbench}, we additionally report
(\romannumeral 4) \textbf{LBE$=\!\frac{E_{\text{vina}}}{N_{\text{lig}}}$ ($\downarrow$)}~\cite{cbgbench}, measuring per-atom binding contribution to reduce size effects.
For the dual-target setting, we further report
(\romannumeral 5) \textbf{Max Vina Dock ($\downarrow$)}, the maximum of the two \textbf{Vina Dock} scores against $P_1$ and $P_2$, and
(\romannumeral 6) \textbf{Dual High Affinity ($\uparrow$)}, the fraction of molecules whose affinities on both targets surpass the reference ligand, reflecting dual-target compatibility~\cite{dualdiff}.

\paragraph{Evaluation protocol.}
We evaluate on the DDF benchmark containing 12,917 dual-target instances, each specifying a pocket pair $(P_1,P_2)$. 
For each instance, we sample $K{=}10$ ligands by each method conditioned on $(P_1,P_2)$, and report the overall mean (Avg.) and median (Med.) over all generated ligands.
For Vina Score, we additionally report the trimmed mean (T-Avg) by discarding the top and bottom 5\% samples to mitigate outliers.

\begin{table}[t]
\caption{\textbf{Molecular property statistics of generated ligands and reference ligands (Avg.).}
($\uparrow$) indicates higher is better.
Top 2 results (excluding Ref.) are highlighted with \textbf{bold text} and \underline{underlined text}, respectively.}
  \label{tab:main_results}
  \begin{center}
    \resizebox{\columnwidth}{!}{
      \begin{tabular}{cccccc}
        \toprule
        \textbf{Methods} &
        \multicolumn{5}{c}{\textbf{Metrics}} \\
        \cmidrule(lr){2-6}
        & QED($\uparrow$) & SA($\uparrow$) & Div.($\uparrow$) & logP & LPSK($\uparrow$) \\
        \midrule
        Ref.              & 0.51           & 0.77           & 0.74           & 2.89 & 4.50 \\\hline
        \textsc{TargetDiff} & 0.54         & 0.55           & 0.67           & 2.32 & 4.69 \\
        \textsc{CompDiff}   & 0.59         & 0.57           & \textbf{0.69}  & 3.48 & 4.74 \\
        \textsc{DualDiff}   & 0.59         & 0.58           & \textbf{0.69}  & 3.54 & \underline{4.88} \\
        \textsc{LinkerNet}  & \underline{0.61} & \underline{0.61} & 0.29       & 3.21 & 4.71 \\
        \textbf{\ours}      & \textbf{0.64} & \textbf{0.68}  & \underline{0.68} & 2.47 & \textbf{4.92} \\
        \bottomrule
      \end{tabular}
      }
  \end{center}

\end{table}

\begin{table*}[h]
\caption{\textbf{Dual-pose quality before docking-based pose search.}
``--'' denotes not applicable (N/A) when a method does not produce target-specific binding poses for the corresponding pocket, and ``OOR'' denotes out-of-range values caused by extreme/invalid poses that fall outside the reported scale.
}
\label{tab:vinascore}
\centering
%\small
% \resizebox{\textwidth}{!}{    
%\resizebox{\columnwidth}{!}{
\begin{tabular}{l ccc ccc cc cc}
\toprule
\multirow{3}{*}{\textbf{Methods}} &
\multicolumn{10}{c}{\textbf{Metrics}} \\
&
\multicolumn{3}{c}{P1-Vina Score ($\downarrow$)} &
\multicolumn{3}{c}{P2-Vina Score ($\downarrow$)} &
\multicolumn{2}{c}{P1-Vina Min ($\downarrow$)} &
\multicolumn{2}{c}{P2-Vina Min ($\downarrow$)} \\
&
Avg. & T-Avg. & Med. &
Avg. & T-Avg. & Med. &
Avg. & Med. &
Avg. & Med. \\
\midrule

Ref. & -7.92 & -7.97 & -7.83
& -7.86 & -7.92 & -7.92
& -8.02 & -7.89
& -7.93 & -7.96 \\\hline
\textsc{TargetDiff} & -6.36 & -7.03 & -7.26 & - & - & - &-7.78 & -7.95& - & - \\
\textsc{CompDiff} & OOR & -1.68 & -3.46 & - & - & - & OOR & -6.26 & - & - \\
\textsc{DualDiff} & OOR & -2.47 & -3.63 & - & - & - & OOR & -6.27 & - & - \\
% \textsc{LinkerNet} & - & - & - & - & - & - & - & - & - & - \\
\textbf{\ours} & -5.99 & -6.39 & -6.39
& -5.81 & -6.35 & -6.42
& -7.47 & -7.57
& -7.50 & -7.52 \\

\bottomrule
\end{tabular}
% }
\end{table*}

\begin{table*}[h]
\caption{\textbf{Docking-based interaction evaluation and dual-target compatibility.}
Top 2 results (excluding Ref.) are highlighted with \textbf{bold text} and \underline{underlined text}, respectively.}
\label{tab:vinadock}
\centering
\small
% \resizebox{\textwidth}{!}{
\begin{tabular}{l ccc ccc cc cc}
\toprule
\multirow{3}{*}{\textbf{Methods}} &
\multicolumn{10}{c}{\textbf{Metrics}} \\
&
\multicolumn{3}{c}{P1-Vina Dock ($\downarrow$)} &
\multicolumn{3}{c}{P2-Vina Dock ($\downarrow$)} &
\multicolumn{2}{c}{Max Vina Dock ($\downarrow$)} &
\multicolumn{2}{c}{Dual High Aff. ($\uparrow$)} \\
&
Avg. & Med. & LBE &
Avg. & Med. & LBE &
Avg. & Med. &
Avg. & Med. \\
\midrule

Ref. & -8.23 & -7.98 & -0.38 & -8.14 & -8.23 & -0.37 & -6.15 & -6.73 & - & - \\\hline
\textsc{TargetDiff} & \underline{-8.95} & \underline{-8.86} & -0.35 & -7.14 & -7.99 & -0.28 & -6.88 & -7.36 & 31.1\% & 23.4\%\\
\textsc{CompDiff} & -8.75 & -8.79 & \underline{-0.36} & \underline{-8.73} & -8.65 & \underline{-0.36} & -7.92 & -7.98 & 39.8\% & 33.0\%\\
\textsc{DualDiff} & -8.80 & -8.82 & \textbf{-0.37} & -8.72 & \underline{-8.78} & \textbf{-0.37} & \underline{-7.99} & \underline{-8.09} & \underline{40.5\%}& \underline{33.4\%}\\
\textsc{LinkerNet} & -8.34 & -8.43 & -0.35 & -8.30 & -8.42 & -0.34 & -7.65 & -7.79 & 38.1\% & 30.5\%\\
\textbf{\ours}
& \textbf{-9.11} & \textbf{-9.13} & \textbf{-0.37}
& \textbf{-9.16} & \textbf{-9.12} & \textbf{-0.37}
& \textbf{-8.37} & \textbf{-8.43} & \textbf{49.2\%} & \textbf{40.9\%}\\
\bottomrule
\end{tabular}
% }
%\vskip -0.1in
\end{table*}

\subsection{Results and Analysis}
Table~\ref{tab:main_results} summarizes molecular property statistics of the generated ligands. \ours{} achieves the best QED, SA, and LPSK among all methods, while achieving diversity on par with the top-performing baselines; ligands generated by \ours{} also have logP within the drug-like range ($[-0.4, 5.6]$).

We also quantify the generation validity of jointly generating two target-specific poses that correspond to the same molecular graph, which is a more stringent criterion in the dual-target setting.
We define \textbf{Dual-Validity} as the percentage of samples for which OpenBabel can assemble a chemically valid molecule from both generated poses given a single atom/bond assignment (i.e., both $(G,X_1)$ and $(G,X_2)$ are valid).
\ours{} achieves \textbf{61\%} Dual-Validity.
This criterion is not applicable to prior methods, as they cannot directly generate a molecule with two target-specific poses.
In contrast, \ours{} directly generates two binding poses and assesses their joint assembly validity, thereby establishing \textbf{Dual-Validity} as a standardized measure for this criterion.

Table~\ref{tab:vinascore} reports pose-dependent \textbf{Vina Score} and \textbf{Vina Min}, which are only defined for methods that explicitly output a pocket-specific binding pose prior to docking-based pose search. We denote \textbf{N/A} (``--'') when such a pose is unavailable (no pose output) or lies outside the Vina scoring box (out-of-box), and use \textbf{OOR} when a pose is available but yields extremely unfavorable scores. \textsc{LinkerNet} is not target-aware and does not generate binding poses, so these scores are not applicable (\textbf{N/A}) for both pockets. \textsc{TargetDiff} generates ligands/poses conditioned on $P_1$, and the resulting conformations are typically out-of-box for $P_2$, making $P_2$ pose-dependent scoring \textbf{N/A}. Similarly, \textsc{DualDiff}/\textsc{CompDiff} sample a single conformation from the composed conditional $p(G,X\mid P_1,\mathcal{T}P_2)$, which concentrates near $P_1$ in the aligned frame; therefore, pose-dependent scoring against the original $P_2$ prior to search is \textbf{N/A}. In contrast, \ours{} directly generates a single molecular graph with two pocket-specific poses $(X_1,X_2)$, establishing the first benchmark for evaluating dual-target pose quality before docking-based search while achieving competitive Vina scores on both pockets.

For docking-based metrics reported in Table~\ref{tab:vinadock}, our method achieves state-of-the-art results on this test set, obtaining the best Vina Dock on both $P_1$ and $P_2$, as well as the lowest Max Vina Dock and the highest Dual High Affinity.
These results indicate that, despite strictly enforcing atom/bond-type consistency and jointly generating two pocket-specific conformations, our model can still produce high-quality dual-target ligands with strong and balanced binding across both targets.

\subsection{Ablation Studies}\label{subsec:ablation}
We conduct ablations on \ours{} targeting two components that are crucial for ensuring \emph{topology consistency} and \emph{symmetry preservation} (Sec.~\ref{overview}): \textit{bond generative modeling} and \textit{DLCF}.
Unlike conventional ablations that lead to gradual performance drops, removing either component breaks the consistency constraints required by the dual-target setting and causes a near-complete collapse in assembly validity.
Detailed ablation setups, failure modes, and quantitative results are deferred to Appendix~\ref{app:ablation_details}.

\begin{figure}[h]
\centering
\includegraphics[width=\linewidth]{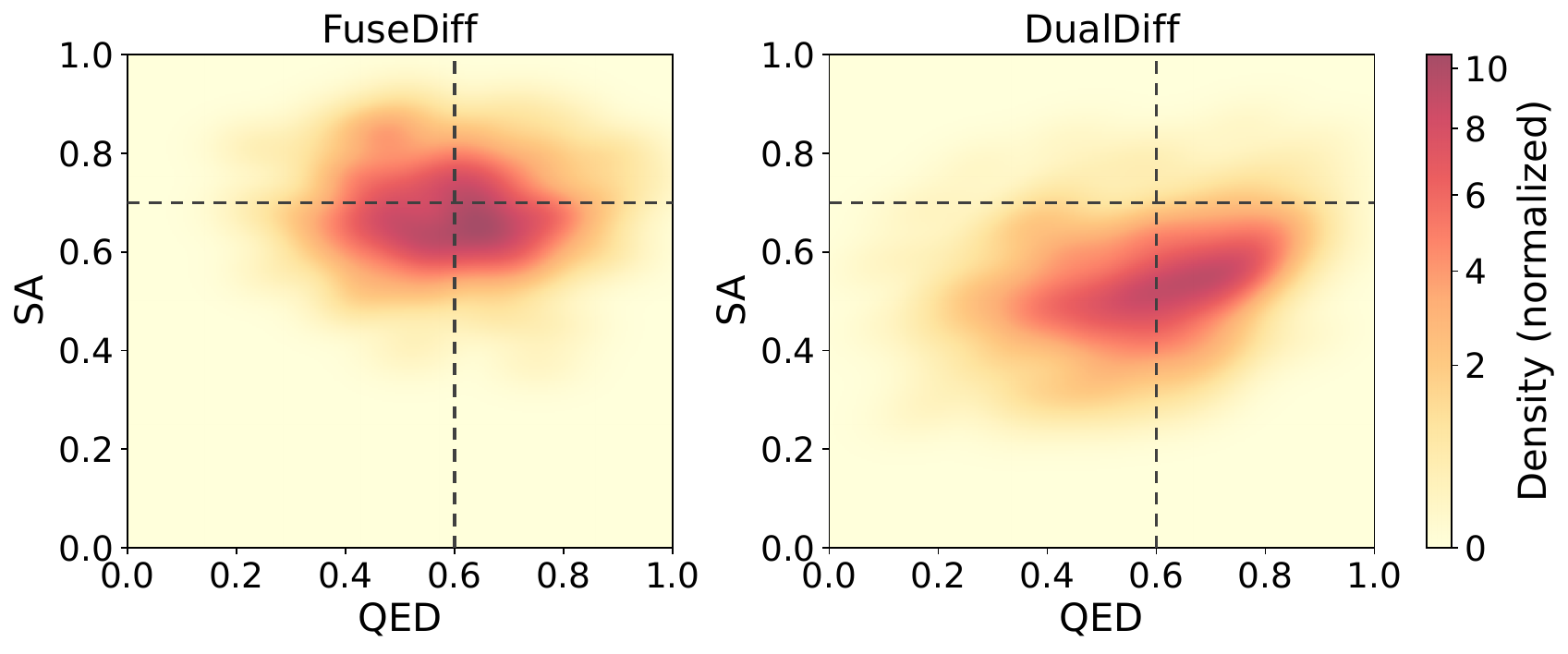}
\caption{\textbf{QED--SA density of filtered molecules ($\mathrm{LPSK}\!=\!5$, $\mathrm{logP}\!\in\![-0.4,5.6]$).}
KDE heatmaps show $\mathrm{QED}$ (x-axis) versus $\mathrm{SA}$ (y-axis) for \ours{} (left) and \textsc{DualDiff} (right).
The upper-right region corresponds to \textbf{DL} ($\mathrm{QED}\!>\!0.6$, $\mathrm{SA}\!>\!0.7$), where \ours{} shows higher density, consistent with its larger DL yield (78 vs.\ 15).}
\label{fig:rw_qed_sa}
\end{figure}

\begin{figure}[h]
\centering
\includegraphics[width=\linewidth]{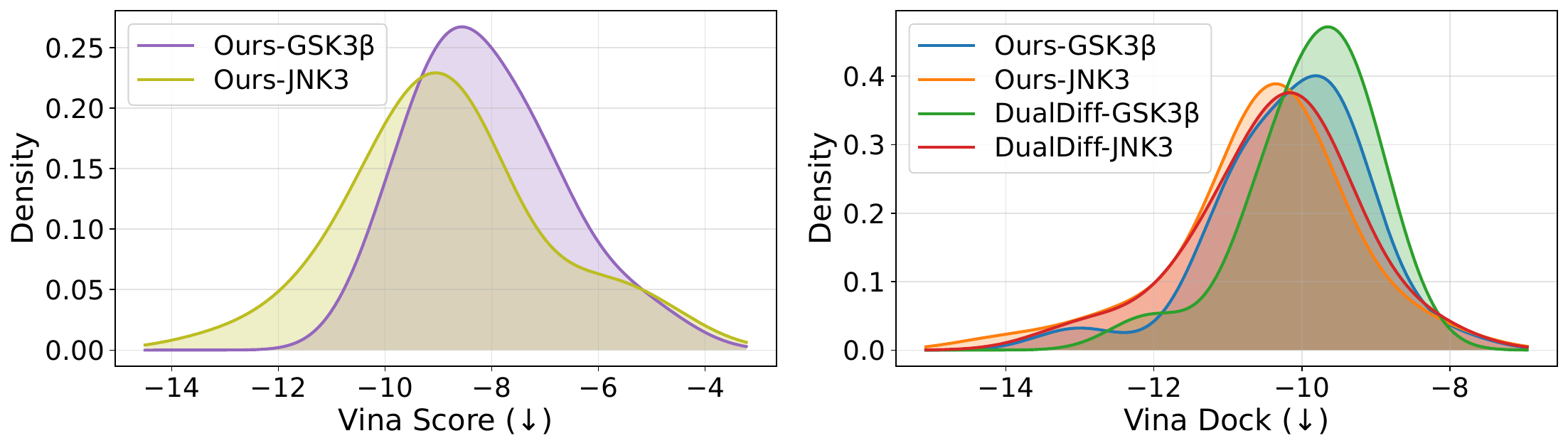}
\caption{\textbf{Vina Score and Vina Dock distributions on GSK3$\beta$ and JNK3.}
\textbf{Left:} \ours{} \textit{Vina Score} on generated poses.
\textbf{Right:} \textit{Vina Dock} after Vina conformational search for \ours{} and \textsc{DualDiff}.}
\label{fig:rw_vina_dist}
\end{figure}

\subsection{Study on Real-world Cases}
\label{subsec:casestudy}
In this section, to evaluate the capability of \ours{} in exploring promising ligands for realistic dual-target drug design scenarios, we establish a dual-target inhibitor design task on two kinases related to Alzheimer's Disease (AD): glycogen synthase kinase-3 beta (GSK3$\beta$, PDB: 6Y9S) and c-Jun N-terminal kinase 3 (JNK3, PDB: 4WHZ). To broaden the real-world validation, we additionally evaluate \ours{} on another therapeutically relevant dual-target pair, ROR$\gamma$t--DHODH, in Appendix~\ref{app:ror-dhodh}.
This target pair has been widely used as a representative real-world test case in prior dual-target molecular design studies\cite{combimots, aixfuse}, and designing inhibitors that simultaneously act on GSK3$\beta$ and JNK3 is regarded as a potential polypharmacology strategy for AD~\cite{LI2023115817}.

\paragraph{Drug-like molecules.}
We sample 1000 molecules for each method and define \textbf{DL} (drug-like) as
$\mathrm{QED}\!>0.6,\mathrm{SA}\!>\!0.7,\mathrm{logP}\!\in\![-0.4,5.6]$, and $\mathrm{LPSK}\!=\!5$.
Under this unified criterion, \ours{} yields substantially more DL molecules than \textsc{DualDiff} (78 vs.\ 15).
Fig.~\ref{fig:rw_qed_sa} shows the QED--SA KDE heatmaps of molecules with $\mathrm{LPSK}\!=\!5$ and $\mathrm{logP}\!\in\![-0.4,5.6]$ generated by \ours{} and \textsc{DualDiff}; \ours{} assigns substantially more probability mass to the DL region ($\mathrm{QED}>0.6$, $\mathrm{SA}>0.7$).

\paragraph{Interaction.}
For the DL molecules generated by \ours{} and \textsc{DualDiff}, we evaluate structure-based binding affinity proxies on both pockets using AutoDock Vina and summarize the score distributions with KDE (Fig.~\ref{fig:rw_vina_dist}). 
Specifically, the left panel reports \textit{Vina Score} of \ours{} evaluated on the generated poses (without conformational optimization or search), while the right panel reports \textit{Vina Dock} after Vina docking search for both methods on both pockets. 
Overall, \ours{} achieves \textit{Vina Score} comparable to the \textit{Vina Dock} of \textsc{DualDiff}, and further yields consistently better \textit{Vina Dock} distributions than \textsc{DualDiff} on both pockets.

\begin{figure}[h]
    \centering
    \setlength{\tabcolsep}{1pt}
    \renewcommand{\arraystretch}{1.0}
    \begin{tabular}{@{}c c c c@{}}
        & \textbf{Reference} & \textbf{\ours{}-1} &  
        \textbf{\ours{}-2}\\
        % \makebox[0.305\linewidth][c]{\shortstack{\textbf{FuseDiff}\\(example 2)}} \\

        \multirow{2}{*}[+8ex]{\rotatebox{90}{\scriptsize\textbf{GSK3$\beta$}}} &
        \includegraphics[width=0.305\linewidth]{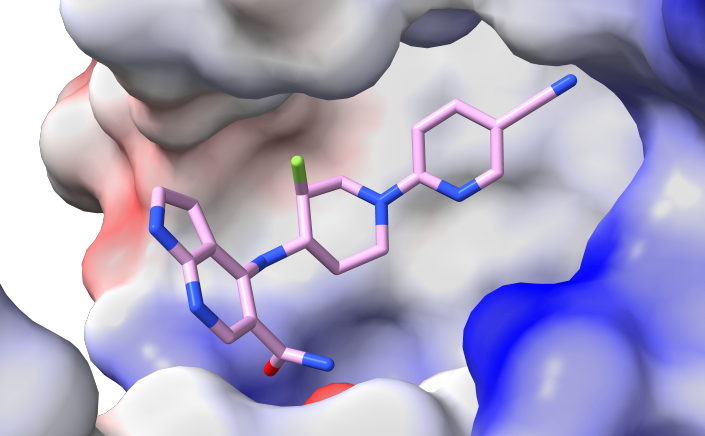} &
        \includegraphics[width=0.305\linewidth]{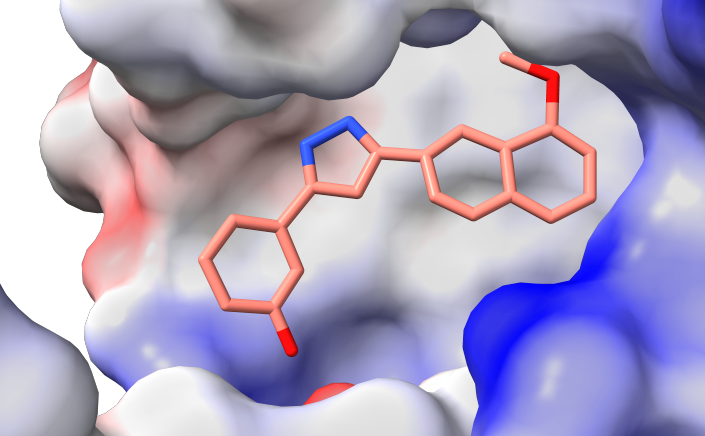} &
        \includegraphics[width=0.305\linewidth]{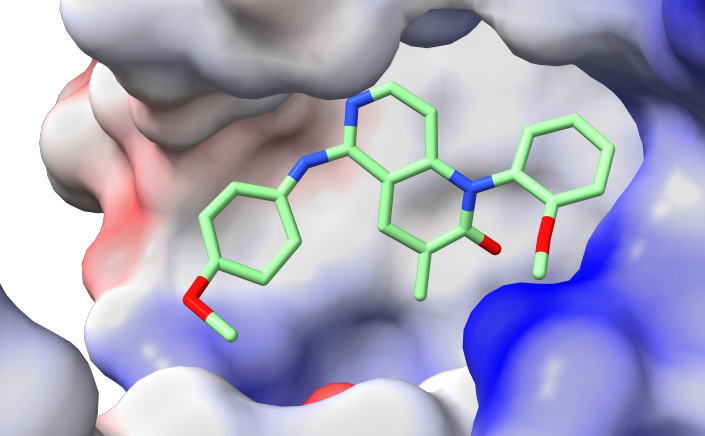} \\
        & {\scriptsize Vina Dock: -9.055} &
          {\scriptsize Vina Score: -8.828} &
          {\scriptsize Vina Score: -9.238} \\[-1pt]

        \multirow{2}{*}[+8ex]{\rotatebox{90}{\scriptsize\textbf{JNK3}}} &
        \includegraphics[width=0.305\linewidth]{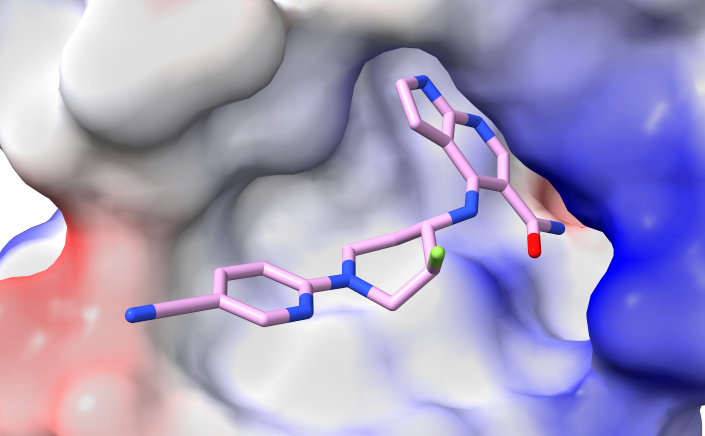} &
        \includegraphics[width=0.305\linewidth]{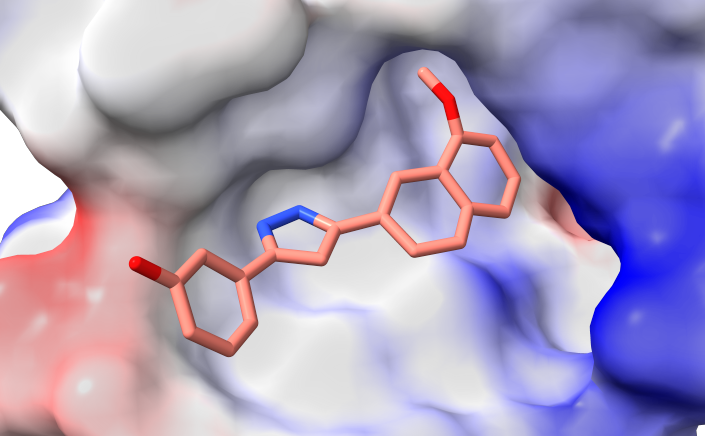} &
        \includegraphics[width=0.305\linewidth]{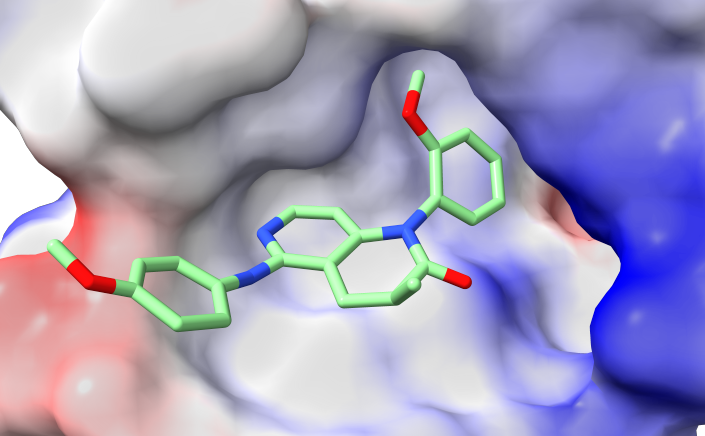} \\
        & {\scriptsize Vina Dock: -9.019} &
          {\scriptsize Vina Score: -8.637} &
          {\scriptsize Vina Score: -9.139} \\
    \end{tabular}
    \caption{\textbf{Qualitative visualization in two pockets.}
Top: GSK3$\beta$; bottom: JNK3.
\textbf{Left:} a reference dual-target ligand with poses obtained by Vina pose search.
\textbf{Middle/Right:} two \ours{} samples.}
    \label{fig:rw_vis}
\end{figure}

\paragraph{Visualization.}\label{casestudy:vis}
We curate single-target active molecules from ChEMBL for each pocket (GSK3$\beta$: 2128; JNK3: 791) and identify 3 graph-isomorphic molecules shared by both sets, which we treat as reference dual-target ligands.
Fig.~\ref{fig:rw_vis} provides a qualitative comparison in the two pockets:
the left column shows a reference ligand with its target-specific binding poses obtained via Vina pose search,
while the middle and right columns show two \ours{} samples, including both the generated ligands and their binding poses sampled by the model without any optimization.

\newpage
\section{Limitations and Ethical Considerations}
\textbf{Limitations.}
Due to the scarcity of experimentally resolved dual-target co-complexes, BN2-DT constructs scalable supervision by pairing two single-target complexes that share the same ligand molecular graph across two pockets. Such pairing is an approximation and may depart from the true dual-target co-binding distribution. Further, for large public structural corpora (e.g., BindingNet v2 and DDF), strictly enforcing non-overlap at the pocket level is inherently difficult, as it depends on similarity criteria and homologous pockets may persist without exact identity, potentially overstating generalization. Progress will benefit from curated dual-target co-complex datasets and stricter split protocols with explicit pocket-similarity controls; our formulation is compatible with such supervision and is expected to benefit as it becomes available.

\noindent\textbf{Ethical Considerations.}
We use only publicly available structural data and benchmarks and comply with their licenses/terms; we will release code and processed data for research use to support reproducibility.
\ours{} outputs computational hypotheses (a ligand graph with two target-specific poses) and should be used only with expert review, safety screening, and experimental validation prior to any real-world application to mitigate misuse.

\section{Conclusion}
To our knowledge, \ours{} is the first symmetry-preserving diffusion model for end-to-end dual-target SBDD that jointly generates a shared ligand graph and its two pocket-specific binding poses. 
\ours{} achieves this by Dual-target Local Context Fusion (DLCF) and explicit bond modeling.
Trained on our constructed BN2-DT dataset, \ours{} enables the first systematic pre-docking evaluation of dual-pose quality and 
achieves state-of-the-art docking performance on the DualDiff benchmark and demonstrates practical utility in real-world case studies.
More broadly, our study suggests a practical design principle for multi-target SBDD with the potential to accelerate polypharmacology-oriented drug discovery.

\section{GenAI Disclosure}
We disclose the use of generative AI tools during the preparation of this manuscript. Specifically, we used a large language model to assist with English polishing and editorial refinement. All technical content—including the method design, experimental setup, results, analysis, and conclusions—was originally written by the authors. The authors reviewed, verified, and substantially revised all AI-assisted edits. The final manuscript was fully edited and approved by the authors.

\begin{acks}
    This research was supported in part by National Natural Science Foundation of China (No. 92470128, No. U2241212, No. 61932001, No. 62302537) and Guangdong Basic and Applied Basic Research Foundation (2025A1515012856).
\end{acks}

\newpage

\bibliographystyle{ACM-Reference-Format}
\balance
\bibliography{sample-base}

\appendix

\section{Dataset~(BN2-DT) Derivation Details}
\label{app:dataset}

\paragraph{Reproducibility.}
Detailed construction scripts, processed data, and implementation details for BN2-DT are provided in our released repository: \url{https://github.com/SYSUWuJL/FuseDiff}. We therefore only summarize the dataset statistics and train--test relationship here.

\paragraph{Statistics and train--test overlap analysis.}
From the BindingNet v2 high confidence subset, we obtain 197,488 ligands in total. Among them, 21,721 ligands appear in at least two targets/pockets and yield 58,058 dual-target training tuples in BN2-DT. After curation and deduplication, BN2-DT contains 801 unique protein targets and 21,721 unique ligands. To assess potential train--test leakage, we measure exact overlap between BN2-DT and the DDF test set at the unordered target-pair level: 332 out of 12,917 DDF test pairs also appear in BN2-DT, corresponding to an exact target-pair overlap rate of 2.57\%. We further use FLAPP~\cite{flapp} to quantify nearest-neighbor pocket similarity between DDF and BN2-DT pairs; the resulting distribution over DDF has mean 0.517, median 0.535, 90th percentile 0.645, and maximum 1.0, suggesting that DDF is not dominated by extremely close training neighbors. At the molecule level, DDF contains 438 unique reference ligands, of which 49 also appear in BN2-DT, corresponding to an exact ligand overlap rate of 11.2\%. These reference ligands are used only for evaluation and are not used as conditioning inputs during generation.

\paragraph{limitation.}
We note that BN2-DT is a derived training set rather than a native collection of experimentally co-observed dual-target binding events. Each tuple is constructed by pairing two independently observed or modeled single-target complexes sharing the same ligand graph. We therefore do not claim that BN2-DT exactly matches the native dual-target binding distribution. Instead, BN2-DT should be viewed as a scalable surrogate for training end-to-end models of $p(G, X_1, X_2 \mid P_1, P_2)$ uunder current data availability.

The proposed \ours{} formulation is independent of this particular data construction and can be directly trained on native dual-target structural datasets when such resources become available.

\section{Theoretical analysis}
\label{app:theory}

\paragraph{Proof of Proposition~\ref{prop:r1r4}.}
We verify \textbf{R1--R4} in Def.~\ref{def:requirements}.

\paragraph{R1 (Permutation invariance).}
Let $\pi$ denote the operator that swaps the two targets:
$\pi(P_1,P_2)=(P_2,P_1)$ and $\pi(G,X_1,X_2)=(G,X_2,X_1)$.
We need to show
$p_\theta(M^{(0)}\mid P)=p_\theta(\pi(M^{(0)})\mid \pi(P))$.

First, the base distribution is symmetric:
$p_{\text{base}}(X_1^{(T)})=p_{\text{base}}(X_2^{(T)})$ and it does not privilege either target,
hence $p_{\text{base}}(M^{(T)}\mid P)=p_{\text{base}}(\pi(M^{(T)})\mid \pi(P))$.

Second, it suffices to show the denoiser $f_\theta(M^{(t)},t,P)$ is permutation-equivariant w.r.t.\ swapping target indices.
In each MPNN layer, \ours{} constructs two point-cloud graphs $G_k^{(\ell)}$ and fuses them into an augmented graph $G_{\text{aug}}^{(\ell)}$.
Swapping $(P_1,P_2)$ exchanges the two point-cloud graphs but leaves the ligand subgraph unchanged.
For ligand--ligand edges, the geometric features are built from the symmetric statistics
$s_{uv}^{(\ell)}=d_{1,uv}^{(\ell)}+d_{2,uv}^{(\ell)}$ and $\delta_{uv}^{(\ell)}=\lvert d_{1,uv}^{(\ell)}-d_{2,uv}^{(\ell)}\rvert$,
which are invariant under swapping the two targets.
All node/edge updates are computed by shared MLPs and permutation-invariant aggregations (sums over neighbors),
therefore swapping the targets permutes the resulting representations in the same way.
Finally, the coordinate updates are carried out separately on each $G_k^{(\ell)}$; after swapping targets, the update for $k=1$ becomes the update for $k=2$ and vice versa.
Hence
$f_\theta(\pi(M^{(t)}),t,\pi(P))=\pi(f_\theta(M^{(t)},t,P))$,
which establishes \textbf{R1}.

\paragraph{R2 (SE(3) invariance).}
We consider independent rigid motions $g,g'\in SE(3)$ applied to the two complexes, i.e.,
$(P_1,X_1)\mapsto (T_gP_1,T_gX_1)$ and $(P_2,X_2)\mapsto (T_{g'}P_2,T_{g'}X_2)$.

(i) The base distributions for $X_1^{(T)}$ and $X_2^{(T)}$ are isotropic Gaussians (restricted to the CoM-free subspace under the standard centering preprocessing used in SBDD diffusion),
hence invariant under rotations and compatible with translation-free (CoM-free) coordinates.

(ii) In message passing, all geometric inputs injected into $G_{\text{aug}}^{(\ell)}$ are scalar distances
(either $d_{k,uv}^{(\ell)}$ for pocket-involved edges or $(s_{uv}^{(\ell)},\delta_{uv}^{(\ell)})$ for ligand--ligand edges),
which are invariant under $SE(3)$ transforms applied to the corresponding complex.
Therefore the updated node/edge embeddings are invariant to rigid motions of either complex.

(iii) The coordinate update in each point-cloud graph $G_k^{(\ell)}$ is of the form (Eq.~(7))
$\Delta r_{k,u}^{(\ell)}=\sum_{w\in N_k(u)} \psi(\cdot)\, \frac{r_{k,u}^{(\ell)}-r_{k,w}^{(\ell)}}{\|r_{k,u}^{(\ell)}-r_{k,w}^{(\ell)}\|_2}$,
i.e., a scalar function of invariant inputs multiplying a relative displacement vector.
Thus, for each $k$, the update is $O(3)$-equivariant for $X_k$ under rigid motions of the $k$-th complex.
Moreover, since the $k$-th coordinate update uses only geometry from $(P_k,X_k)$, it is invariant to independent rigid motions applied to the other complex.
This establishes \textbf{R2}.

\paragraph{R3 (Non-factorization).}
Although each reverse transition factorizes as
$p_\theta(M^{(t-1)}\mid M^{(t)},P)=\prod_{k=1}^2 p_\theta(X_k^{(t-1)}\mid M^{(t)},P)\cdot p_\theta(G^{(t-1)}\mid M^{(t)},P)$,
the model density is obtained by marginalizing the latent trajectory:
\[
p_\theta(M^{(0)}\mid P)
=
\int p_{\text{base}}(M^{(T)}\mid P)\prod_{t=1}^{T}p_\theta(M^{(t-1)}\mid M^{(t)},P)\, dM^{(1:T)}.
\]
Crucially, both $X_1^{(t-1)}$ and $X_2^{(t-1)}$ are conditioned on the shared latent state
$M^{(t)}=(G^{(t)},X_1^{(t)},X_2^{(t)})$ which is integrated out.
In graphical-model terms, $X_1^{(t-1)} \leftarrow M^{(t)} \rightarrow X_2^{(t-1)}$ makes them d-connected after marginalizing $M^{(t)}$,
so the resulting joint need not factorize as $p(X_1,X_2\mid G,P_1,P_2)=p(X_1\mid G,P_1,P_2)\,p(X_2\mid G,P_1,P_2)$.
Therefore $X_1 \not\!\perp X_2 \mid (P_1,P_2,G)$, establishing \textbf{R3}.

\paragraph{R4 (Non-restricted support).}
\ours{} never imposes a global rigid alignment constraint between the two target-specific complexes.
While DLCF couples the two targets through shared ligand embeddings computed on the fused augmented graph,
the coordinate updates are performed \emph{separately} within each point-cloud graph $G_k^{(\ell)}$.
The displacement field for target $k$ depends on target-specific geometric inputs such as the neighbor set $N_k(u)$ and distances induced by $P_k$.
Since $P_1\neq P_2$ in general, these inputs can vary independently across targets, allowing the sampled $(X_1,X_2)$ to vary beyond a single shared rigid transform.
Hence the support is not restricted to $\{(X_1,X_2): \exists g\in SE(3)\ \text{s.t.}\ X_1=T_gX_2\}$, establishing \textbf{R4}.
\qed

\section{Single-target Pretraining Checkpoint for Fair Benchmarking}
\label{app:targetdiff_bn2}

To control for differences in single-target pretraining data when benchmarking \textsc{TargetDiff}, \textsc{DualDiff}, and \textsc{CompDiff} on DDF, we compare two pretrained \textsc{TargetDiff} checkpoints: the official one pretrained on CrossDocked2020 (denoted as \textsc{-CD}) and our checkpoint pretrained on BindingNet v2 (denoted as \textsc{-BN2}).
All methods are evaluated on the DDF benchmark with the same sampling and evaluation protocol.
As shown in Tables~\ref{tab:ckpt_props}--\ref{tab:ckpt_dock}, the two checkpoints lead to overall comparable performance across both chemical-property metrics and docking-based affinity metrics, indicating that the downstream comparisons are not driven by the particular choice of single-target pretraining data.
We therefore use the \textsc{-BN2} checkpoint as a common pretrained backbone in our main experiments for a controlled comparison.

\begin{table}[h]
\caption{\textbf{Checkpoint comparison on DDF (molecular properties).}
We report molecule-level metrics for \textsc{TargetDiff}, \textsc{CompDiff}, and \textsc{DualDiff} when using the official CrossDocked2020 pretrained checkpoint (\textsc{-CD}) versus our BindingNet v2 pretrained checkpoint (\textsc{-BN2}).}
\label{tab:ckpt_props}
  \begin{center}
    \resizebox{\columnwidth}{!}{
      \begin{tabular}{cccccc}
        \toprule
        \textbf{Methods} &
        \multicolumn{5}{c}{\textbf{Metrics}} \\
        \cmidrule(lr){2-6}
        & QED($\uparrow$) & SA($\uparrow$) & Div.($\uparrow$) & logP & LPSK($\uparrow$) \\
        \midrule
        Ref.              & 0.51           & 0.77           & 0.74           & 2.89 & 4.50 \\\hline
        \textsc{TargetDiff}-CD & 0.53&0.54 &0.70 &2.48 &4.63 \\
        \textsc{TargetDiff}-BN2 & 0.54         & 0.55           & 0.67           & 2.32 & 4.69 \\
        \textsc{CompDiff}-CD & 0.57 & 0.58 & 0.72 & 3.67 & 4.54 \\
        \textsc{CompDiff}-BN2 & 0.59 & 0.57 & 0.69 & 3.48 & 4.74 \\
        \textsc{DualDiff}-CD & 0.59 & 0.61 & 0.67& 3.43& 4.67\\
        \textsc{DualDiff}-BN2  & 0.59 & 0.58 & 0.69 & 3.54 & 4.88 \\
        \bottomrule
      \end{tabular}
      }
  \end{center}
\end{table}

\begin{table*}[!b]
\caption{\textbf{Checkpoint comparison on DDF (docking-based affinity).}
Docking metrics on the two pockets (P1/P2) and their aggregate (Max Vina Dock), together with Dual High Affinity, for baselines using \textsc{-CD} versus \textsc{-BN2} pretrained \textsc{TargetDiff} checkpoints.}
\label{tab:ckpt_dock}
\centering
\small
% \resizebox{\linewidth}{!}{
\begin{tabular}{l ccc ccc cc cc}
\toprule
\multirow{3}{*}{\textbf{Methods}} &
\multicolumn{10}{c}{\textbf{Metrics}} \\
&
\multicolumn{3}{c}{P1-Vina Dock ($\downarrow$)} &
\multicolumn{3}{c}{P2-Vina Dock ($\downarrow$)} &
\multicolumn{2}{c}{Max Vina Dock ($\downarrow$)} &
\multicolumn{2}{c}{Dual High Aff. ($\uparrow$)} \\
&
Avg. & Med. & LBE &
Avg. & Med. & LBE &
Avg. & Med. &
Avg. & Med. \\
\midrule

Ref. & -8.23 & -7.98 & -0.38 & -8.14 & -8.23 & -0.37 & -6.15 & -6.73 & - & - \\\hline
\textsc{TargetDiff}-CD & -8.89& -8.91 &-0.37 & -7.23 & -7.68& -0.28& -6.84& -6.91& 30.8\%& 23.3\% \\
\textsc{TargetDiff}-BN2 & -8.95 & -8.86 & -0.35 & -7.14 & -7.99 & -0.28 & -6.88 & -7.36 & 31.1\% & 23.4\%\\
\textsc{CompDiff}-CD & -8.35 & -8.46 & -0.36 & -8.35 & -8.37 & -0.36 & -7.53 & -7.84 & 37.9\% & 31.4\% \\
\textsc{CompDiff}-BN2 & -8.75 & -8.79 & -0.36 & -8.73 & -8.65 & -0.36 & -7.92 & -7.98 & 39.8\% & 33.0\%\\
\textsc{DualDiff}-CD & -8.46 & -8.52 & -0.38 & -8.45 & -8.52 & -0.38 & -7.77 & -7.79 & 39.3\% & 31.9\% \\
\textsc{DualDiff}-BN2 & -8.80 & -8.82 & -0.37 & -8.72 & -8.78 & -0.37 & -7.99 & -8.09 & 40.5\% & 33.4\%\\
\bottomrule
\end{tabular}
% }
%\vskip -0.1in
\end{table*}

\section{Ablation Details}\label{app:ablation_details}

\paragraph{(1) W/o bond generative modeling}
We ablate explicit bond-type generation by diffusing and sampling only atom types and coordinates, while still performing message passing and representation updates with DLCF on the augmented graph.
Instead of predicting bond types from augmented edge representations via a dedicated bond head as in \ours{}, we infer chemical bonds \emph{post hoc} and \emph{independently} for each pose from the shared atom types and the denoised coordinates.
This decoupled reconstruction frequently yields pose-specific bond assignments that are mutually incompatible across $X_1$ and $X_2$, preventing the recovery of a single molecular graph consistent with both targets.
As a result, \textbf{Dual-Validity} drops from 61\% in the full model to 3\% on DDF, indicating that generative bond modeling is essential for enforcing cross-pose topological consistency.

\paragraph{(2) W/o DLCF}
We ablate Dual-target Local Context Fusion (DLCF) and thus do not fuse the two point-cloud graphs (Sec.~\ref{subsubsec:graphsetup}) into a single augmented graph.
To preserve topology consistency, a natural fallback is to sample sequentially under the two pockets while enforcing a shared ligand graph, resembling a straightforward extension of single-target SBDD.
However, this strategy effectively marginalizes over the second pocket during the first-stage generation, making cross-target incompatibility systematic rather than incidental.
Empirically, the resulting samples exhibit pronounced atom--bond inconsistencies (e.g., unphysical covalent distances and distorted conformations), deviating from realistic molecular geometries.
Consequently, \textbf{Dual-Validity} drops to 2\%, demonstrating that DLCF is necessary to realize joint conditionality (R3) without sacrificing geometric plausibility.

\section{Additional Real-world Case: ROR$\gamma$t--DHODH}
\label{app:ror-dhodh}

To broaden the real-world validation beyond the GSK3$\beta$--JNK3 case in Sec.~\ref{subsec:casestudy}, we additionally evaluate \ours{} on another therapeutically relevant dual-target pair, ROR$\gamma$t--DHODH. We follow the same generation and filtering protocol as in Sec.~\ref{subsec:casestudy}, and apply the same drug-like criterion, i.e., QED $>$ 0.6, SA $>$ 0.7, logP $\in [-0.4, 5.6]$, and LPSK $= 5$. Under this criterion, \ours{} again produces substantially more drug-like molecules than DualDiff, yielding 41 candidates compared with 5 from DualDiff, and its QED--SA distribution shifts toward a more favorable high-QED and high-SA region. These results suggest that the empirical behavior of \ours{} is not limited to a single real-world target pair, providing additional evidence for its applicability to realistic dual-target design scenarios.

\end{document}